\documentclass[a4paper,twocolumn,english]{paper}
\usepackage[T1]{fontenc}
\usepackage[latin9]{inputenc}
\usepackage{xcolor}
\usepackage{babel}
\usepackage{array}
\usepackage{units}
\usepackage{multirow}
\usepackage{amsmath}
\usepackage{amssymb}
\usepackage{graphicx}
\usepackage[unicode=true]
 {hyperref}

\makeatletter

\pdfpageheight\paperheight
\pdfpagewidth\paperwidth

\providecommand{\tabularnewline}{\\}

\usepackage{xcolor}

\usepackage{url}

\definecolor{tab_red}{RGB}{231,0,34}
\definecolor{tab_orange}{RGB}{255,114,4}
\definecolor{tab_olive}{RGB}{176,211,41}
\definecolor{tab_green}{RGB}{0,188,50}
\definecolor{tab_blue}{RGB}{35,100,180}

\definecolor{tab_grey}{RGB}{75,75,75}

\definecolor{initial}{RGB}{11,36,251}
\definecolor{iutisfive}{RGB}{15,127,18}
\definecolor{cnnsfc}{RGB}{252,13,27}
\definecolor{twentysix}{RGB}{128,128,128}
\definecolor{maxnine}{RGB}{204,204,204}

\usepackage{mathtools}
\usepackage{pifont}

\usepackage{xspace}
\newcommand{\AllCombinations}{\emph{All Combinations}\xspace}
\newcommand{\RandomChoice}{\emph{Random Choice}\xspace}
\newcommand{\PropFG}{\emph{Prop. FG}\xspace}
\newcommand{\AveragedBayes}{\emph{Averaged Bayes}\xspace}
\newcommand{\BKS}{\emph{BKS}\xspace}
\newcommand{\MajorityVote}{\emph{Majority Vote}\xspace}
\newcommand{\CDNET}{CDnet\xspace}

\makeatother

\begin{document}
\global\long\def\initial{{\color{initial}\bullet}}%
\global\long\def\iutisfive{{\color{iutisfive}\bullet}}%
\global\long\def\cnnsfc{{\color{cnnsfc}\bullet}}%
\global\long\def\randomselection{\square}%
\global\long\def\maxnine{\randomselection\cup{\color{maxnine}\blacksquare}}%
\global\long\def\twentysix{\maxnine\cup{\color{twentysix}\blacksquare}}%

\global\long\def\BackgroundSubtraction{\text{BGS}}%

\global\long\def\Background{\text{BG}}%

\global\long\def\Foreground{\text{FG}}%

\global\long\def\etal{\mathit{et\,al.}}%

\global\long\def\etc{\mathit{etc}}%

\global\long\def\ie{\mathit{i.e.}}%

\global\long\def\eg{\mathit{e.g.}}%

\global\long\def\threshold{\tau}%

\global\long\def\thresholdingFunction#1{\mathbf{1}_{\ge#1}}%



\global\long\def\comma{\enspace\mbox{,}}%

\global\long\def\dot{\enspace\mbox{.}}%


\global\long\def\PAWCS{\text{{\color{teal}PAWCS}}}%
\global\long\def\SuBSENSE{\text{{\color{teal}SuBSENSE}}}%
\global\long\def\WeSamBE{\text{{\color{teal}WeSamBE}}}%
\global\long\def\SharedModel{\text{{\color{teal}SharedModel}}}%
\global\long\def\FTSG{\text{{\color{teal}FTSG}}}%
\global\long\def\CwisarDRP{\text{{\color{teal}CwisarDRP}}}%
\global\long\def\MBS{\text{{\color{teal}MBS}}}%
\global\long\def\CEFIC{\text{{\color{teal}CEFIC}}}%
\global\long\def\MBSv{\text{{\color{teal}MBSv0}}}%

\global\long\def\CwisarDH{\text{{\color{blue}CwisarDH}}}%
\global\long\def\EFIC{\text{{\color{blue}EFIC}}}%
\global\long\def\Spectral{\text{{\color{blue}Spectral360}}}%
\global\long\def\BMOG{\text{{\color{blue}BMOG}}}%
\global\long\def\AMBER{\text{{\color{blue}AMBER}}}%
\global\long\def\AAPSA{\text{{\color{blue}AAPSA}}}%
\global\long\def\GraphCutDiff{\text{{\color{blue}GraphCutDiff}}}%
\global\long\def\SCSOBS{\text{{\color{blue}SC\_SOBS}}}%
\global\long\def\RMoG{\text{{\color{blue}RMoG}}}%

\global\long\def\Mahalanobis{\text{{\color{red}Mahalanobis}}}%
\global\long\def\KDE{\text{{\color{red}KDE}}}%
\global\long\def\CPonline{\text{{\color{red}CP3-online}}}%
\global\long\def\GMMStauffer{\text{{\color{red}GMM-Stauffer}}}%
\global\long\def\GMMZivkovic{\text{{\color{red}GMM-Zivkovic}}}%
\global\long\def\SimplifiedOBS{\text{{\color{red}Simplified\_OBS}}}%
\global\long\def\Multiscale{\text{{\color{red}Multiscale}}}%
\global\long\def\Euclidean{\text{{\color{red}Euclidean}}}%

\title{An exploration of the performances achievable by combining unsupervised
background subtraction algorithms}
\author{Sébastien Piérard, Marc Braham, and Marc Van Droogenbroeck}
\maketitle
\begin{abstract}
Background subtraction (BGS) is a common choice for performing motion
detection in video. Hundreds of BGS algorithms are released every
year, but combining them to detect motion remains largely unexplored.
We found that combination strategies allow to capitalize on this massive
amount of available BGS algorithms, and offer significant space for
performance improvement. In this paper, we explore sets of performances
achievable by 6 strategies combining, pixelwise, the outputs of $26$
unsupervised $\BackgroundSubtraction$ algorithms, on the \CDNET
2014 dataset, both in the ROC space and in terms of the F1 score.
The chosen strategies are representative for a large panel of strategies,
including both deterministic and non-deterministic ones, voting and
learning. In our experiments, we compare our results with the state-of-the-art
combinations IUTIS-5 and CNN-SFC, and report six conclusions, among
which the existence of an important gap between the performances of
the individual algorithms and the best performances achievable by
combining them.
\end{abstract}

\section{Introduction}

\label{sec:Introduction}

Background subtraction ($\BackgroundSubtraction$) aims at detecting
pixels belonging to moving objects in video sequences. It has been
very popular over the last decade, and has given rise to a massive
amount of algorithms predicting either the label $\Background=0$
(background) or $\Foreground=1$ (foreground) in each pixel.

Today, the $\BackgroundSubtraction$ community is still working hard
to find ways to push the performance. An overview of the current status
is provided by the \href{http://changedetection.net/}{changedetection.net}
platform. It provides the \CDNET 2014~\cite{Wang2014AnExpanded}
dataset with $53$ reference videos, grouped in $11$ categories,
for a total number of $150,000$ frames annotated manually at the
pixel level. It also makes publicly available the binary outputs of
various algorithms. And, last but not least, it helps in comparing
algorithms, by reporting performance indicators (such as the error
rate $\textrm{ER}$, the true positive rate $\textrm{TPR}$, the false
positive rate $\textrm{FPR}$, the $\textrm{F}_{1}$ score, etc.),
and offers an up-to-date ranking.

Currently, the effort is almost exclusively focussing on the development
of new algorithms, with hundreds of them being designed every year.
Their principles can be found in the surveys~\cite{Bouwmans2014Traditional,Bouwmans2019Deep,Bouwmans2020Background,Mandal2021AnEmpirical}.
Despite the importance of the effort put in this path, the performance
reported on \CDNET is saturating.

An alternative path consists in combining algorithms~\cite{Jodoin2014Overview}.
Surprisingly, only a few papers took this path. The current state-of-the-art
combinations are IUTIS-5~\cite{Bianco2017Combination} and CNN-SFC~\cite{Zeng2018Combining},
which have been obtained by learning.

In this paper, we are also considering the combination of $\BackgroundSubtraction$
algorithms. Our contributions are the following.

First, we innovate by expressing the set of all performances achievable
by combination, rather than discussing a unique algorithm. More precisely,
we explore the pixelwise combinations of $26$ unsupervised $\BackgroundSubtraction$
algorithms, with $6$ combination strategies. We also innovate by
deliberately focussing on the combination of the outputs, instead
of the intrinsic mechanisms for dealing with the input pixel values.

Second, we point out that the \CDNET 2014 platform remains largely
underexploited, and show that the availability of $\BackgroundSubtraction$
algorithm segmentation masks makes it possible to go beyond the production
of a leaderboard. Actually, the evaluated algorithms only represent
a negligible proportion of the possible algorithms. But \CDNET 2014
contains all the necessary information to perform a kind of ``algorithmic
augmentation'' by combining segmentation outputs. By doing so, we
capitalize on the results accumulated over these years.

Third, we demonstrate the richness of our approach. We report the
sets of achievable performances, for all considered combination strategies,
in terms of the Receiver Operating Characteristic $\textrm{ROC}=(\textrm{FPR},\textrm{TPR})$
space. We also provide additional experimental results related to
the $\textrm{F}_{1}$ score. Based on our results, we draw six conclusions.

The outline of this paper is as follows. Section~\ref{sec:Methodology}
details our methodology. Section~\ref{sec:Implementation} gives
an overview of our implementation. We present and analyze results
in Section~\ref{sec:Results-and-observations}. Finally, Section~\ref{sec:Conclusion}
briefly concludes the paper.

\section{Exploration methodology}

\label{sec:Methodology}

Our exploration methodology is built upon the following terms, further
discussed in the subsections: (1) what we combine, (2) how the combinations
are performed, and (3) how the performance is measured.

\subsection{The combined algorithms}

\begin{table}[t]
\resizebox{\columnwidth}{!}{
\begin{tabular}{|c|c|}
\hline 
Rank & Algorithm\tabularnewline
\hline 
1 & $\PAWCS$~\cite{StCharles2016Universal}\tabularnewline
\hline 
2 & $\SuBSENSE$~\cite{StCharles2015SuBSENSE}\tabularnewline
\hline 
3 & $\WeSamBE$~\cite{Jiang2018WeSamBE}\tabularnewline
\hline 
4 & $\SharedModel$~\cite{Chen2015Learning}\tabularnewline
\hline 
5 & $\FTSG$~\cite{Wang2014Static}\tabularnewline
\hline 
6 & $\CwisarDRP$~\cite{DeGregorio2017WiSARD}\tabularnewline
\hline 
7 & $\MBS$~\cite{Sajid2017Universal}\tabularnewline
\hline 
8 & $\CEFIC$~\cite{Allebosch2015CEFIC}\tabularnewline
\hline 
9 & $\MBSv$~\cite{Sajid2015Background}\tabularnewline
\hline 
\end{tabular}~%
\begin{tabular}{|c|c|}
\hline 
Rank & Algorithm\tabularnewline
\hline 
10 & $\CwisarDH$~\cite{DeGregorio2014Change}\tabularnewline
\hline 
11 & $\EFIC$~\cite{Allebosch2015EFIC}\tabularnewline
\hline 
12 & $\Spectral$~\cite{Sedky2014Spectral}\tabularnewline
\hline 
13 & $\BMOG$~\cite{Martins2017BMOG}\tabularnewline
\hline 
14 & $\AMBER$~\cite{Wang2014AFastSelfTuning}\tabularnewline
\hline 
15 & $\AAPSA$~\cite{Ramirez-Alonso2016Auto-adaptive}\tabularnewline
\hline 
16 & $\GraphCutDiff$~\cite{Miron2015Change}\tabularnewline
\hline 
17 & $\SCSOBS$~\cite{Maddalena2012TheSOBS}\tabularnewline
\hline 
18 & $\RMoG$~\cite{Varadarajan2013Spatial}\tabularnewline
\hline 
\end{tabular}~%
\begin{tabular}{|c|c|}
\hline 
Rank & Algorithm\tabularnewline
\hline 
19 & $\Mahalanobis$~\cite{Benezeth2010Comparative}\tabularnewline
\hline 
20 & $\KDE$~\cite{Elgammal2000NonParametric}\tabularnewline
\hline 
21 & $\CPonline$~\cite{Liang2015Cooccurence}\tabularnewline
\hline 
22 & $\GMMStauffer$~\cite{Stauffer1999Adaptive}\tabularnewline
\hline 
23 & $\GMMZivkovic$~\cite{Zivkovic2004Improved}\tabularnewline
\hline 
24 & $\SimplifiedOBS$~\cite{Sehairi2017Comparative}\tabularnewline
\hline 
25 & $\Multiscale$~\cite{Lu2014AMultiscale}\tabularnewline
\hline 
26 & $\Euclidean$~\cite{Benezeth2010Comparative}\tabularnewline
\hline 
\textbackslash{} & \textbackslash\tabularnewline
\hline 
\end{tabular}}

\caption{Set of unsupervised $\protect\BackgroundSubtraction$ algorithms to
be combined.\label{tab:Pool-of-combinable}}
\end{table}
 We have chosen a set of $\BackgroundSubtraction$ algorithms for
which the binary segmentation masks (outputs) are publicly available
on the \CDNET platform. They are listed in Table~\ref{tab:Pool-of-combinable},
with their relative ranks in the leaderboard. Despite that some of
these $26$ unsupervised algorithms use random numbers, we consider
them as deterministic as only one output is uploaded on the platform.
We run experiments in which the $26$ algorithms are combined, and
others in which number of combined algorithms is limited to $9$,
which is more realistic in practice.

\subsection{The combination strategies}

We have chosen the following strategies to combine the outputs of
the chosen algorithms at the pixel level.

\textbf{\AllCombinations.} In a stochastic perspective, the behavior
of any combiner is given by the probabilities of predicting $\Foreground$
for each of the $2^{n}$ possible joint outputs for the $n$ combined
algorithms. Thus, any combiner can be seen as a point of the $[0,1]^{2^{n}}$
hypercube, and the $2^{2^{n}}$ deterministic combiners $\left\{ 0,1\right\} ^{n}\rightarrow\left\{ 0,1\right\} $
can be seen as its vertices $\{0,1\}^{2^{n}}$.

\textbf{\RandomChoice. }A subset of combinations can be obtained
by choosing, at random and according to fixed probabilities, either
$\Background$, $\Foreground$, or one of the combined outputs.

\textbf{\emph{Deterministic combinations}}\textbf{.} Some deterministic
combinations can be obtained by thresholding ``soft combinations''
whose output is a confidence. Examples include the proportion of algorithms
predicting $\Foreground$~\cite{Jodoin2014Overview} (\PropFG),
the \AveragedBayes's classifier~\cite{Jodoin2014Overview}, and
\BKS~\cite{Huang1995AMethod}. To the best of our knowledge, \BKS
has never been applied to $\BackgroundSubtraction$ algorithms. The
\MajorityVote, defined for any odd $n$, is a particular case of
\PropFG with the threshold value $\threshold=\nicefrac{1}{2}$. Note
that there is no guarantee to improve the performance by the majority
vote~\cite{Kuncheva2003Limits}. The formulas for these four strategies
are given in Table~\ref{tab:combination-strategies}, with the respective
number of distinct combinations that can be obtained by tuning $\threshold$.
\begin{table}
\begin{centering}
\resizebox{\columnwidth}{!}{
\begin{tabular}{|c|c|c|}
\hline 
strategy & combination formula & amount\tabularnewline
\hline 
\hline 
\MajorityVote & $\thresholdingFunction{0.5}\left(\frac{1}{n}\sum_{j=1}^{n}\textrm{b}_{j}\right)$ & $=1$\tabularnewline
\hline 
\PropFG & $\thresholdingFunction{\threshold}\left(\frac{1}{n}\sum_{j=1}^{n}\textrm{b}_{j}\right)$ & $=n+2$\tabularnewline
\hline 
\AveragedBayes & $\thresholdingFunction{\threshold}\left(\frac{1}{n}\sum_{j=1}^{n}\left(\left(1-\textrm{b}_{j}\right)FOR_{j}^{(LS)}+b_{j}PPV_{j}^{(LS)}\right)\right)$ & $\le2^{n}+1$\tabularnewline
\hline 
\BKS & $\thresholdingFunction{\threshold}\left(P_{FG}^{(LS)}\left(\textrm{b}_{1},\ldots,\textrm{b}_{n}\right)\right)$ & $\le2^{n}+1$\tabularnewline
\hline 
\end{tabular}}
\par\end{centering}
\caption{Some deterministic combination strategies. Here, $\protect\thresholdingFunction{\protect\threshold}$
denotes a thresholding operation w.r.t. a threshold $\protect\threshold$.
The output of the $j$th combined algorithm is $\textrm{b}_{j}\in\left\{ 0,1\right\} $.
The constants $FOR_{j}^{(LS)}$, $PPV_{j}^{(LS)}$, and $P_{FG}^{(LS)}\left(\textrm{b}_{1},\ldots,\textrm{b}_{n}\right)$
are learned with the learning set $LS$ and correspond, respectively,
to the empirical false omission rate of the $j$th combined algorithm,
to its empirical positive predictive value (precision), and to the
probability of foreground given all outputs.\label{tab:combination-strategies}}
\end{table}

Implementing \AveragedBayes requires the knowledge of the precision
and false omission rate of all combined algorithms. For \BKS, we
need to know the probability of foreground for all the possible joint
outputs of the combined algorithms. These quantities are estimated
empirically from a learning set. In order for our results to be comparable
with IUTIS-5 and CNN-SFC, we used the same learning set $LS$ as in
those papers. It is obtained by aggregating all pixels from the shortest
video in each category of \CDNET. This learning set has more than
a billion training samples, which is enough to estimate the quantities
needed by \AveragedBayes and \BKS.

\subsection{Measuring performances}

We determine the performances of the combinations with all the $53$
videos of \CDNET 2014. All categories are equally important, and
all videos within any given category receive also an equal importance.
\CDNET reports the weighted arithmetic mean of the performance indicators
obtained for each video. Another technique, known as \emph{summarization},
presents some advantages~\cite{Pierard2020Summarizing}. Nevertheless,
in this paper, we stick to the technique of \CDNET to calculate $\overline{\textrm{ROC}}=(\overline{\textrm{TPR}},\overline{\textrm{FPR}})$
and $\overline{\textrm{F}_{1}}$, as it is the common practice in
the $\BackgroundSubtraction$ community.

\section{Implementation overview}

\label{sec:Implementation}

\subsection{With the strategy \RandomChoice\label{subsec:Technique-with-the-strategy-random-choice}}

In the $\overline{\textrm{ROC}}$ space, the set of performances achievable
by choosing one algorithm at random corresponds to the convex hull
of the individual performances. In particular, for $n=2$, it corresponds
to the line segment between the individual performances. Note that
a similar property is known in the classical (unweighted) $\textrm{ROC}$
space~\cite{Fawcett2006AnIntroduction}.

\subsection{With the strategy \AllCombinations\label{subsec:Technique-with-the-strategy-all-combinations}}

Any given combination can be expressed as a random choice between
some ($2^{n}+1$ are enough) deterministic combinations. Thus, the
set of all performances achievable by combining the outputs of $n$
algorithms is, in $\overline{\textrm{ROC}}$, the convex hull of the
performances achievable with the $2^{2^{n}}$ deterministic combinations.
When $n$ is large, measuring the performances of all the $2^{2^{n}}$
deterministic combinations is unrealistic ($2^{2^{n}}=1.0938\times10^{20201781}$
with $n=26$). But, as $\overline{\textrm{TPR}}$ and $\overline{\textrm{FPR}}$
are linear with respect to the probabilities to predict $\Foreground$
for the $2^{n}$ possible joint outputs, the achievable area in $\overline{\textrm{ROC}}$
is a linear projection of the hypercube $\left[0,1\right]^{2^{n}}$,
that is a zonotope. We discovered an efficient way to compute the
vertices on its contour, making it possible to compute the set of
achievable performances for large values of $n$ (even for $n=26$!).
For selections involving fewer $\text{\ensuremath{\BackgroundSubtraction}}$
algorithms, we obtain an achievable zonotope per selection and compute
the contour of the union of all these zonotopes.

\subsection{With the other strategies\label{subsec:Technique-with-other-strategies}}

With the other strategies, we proceed by testing each combination
exhaustively, with an optimized software. Note that there are 5 millions
possible selections of $n\le9$ algorithms out of $26$. Just to illustrate
how difficult it has been to explore their combinations, the number
of possible combinations is $5.6585\times10^{6}$ for the \MajorityVote,
$4.7002\times10^{7}$ for \PropFG, and $2.0954\times10^{9}$ for
\AveragedBayes and \BKS. In addition, each combination requires
to read 12 billions pixels.

\section{Results and observations}

\label{sec:Results-and-observations}

\begin{figure}[t]
\centering{}\includegraphics[width=0.9\columnwidth]{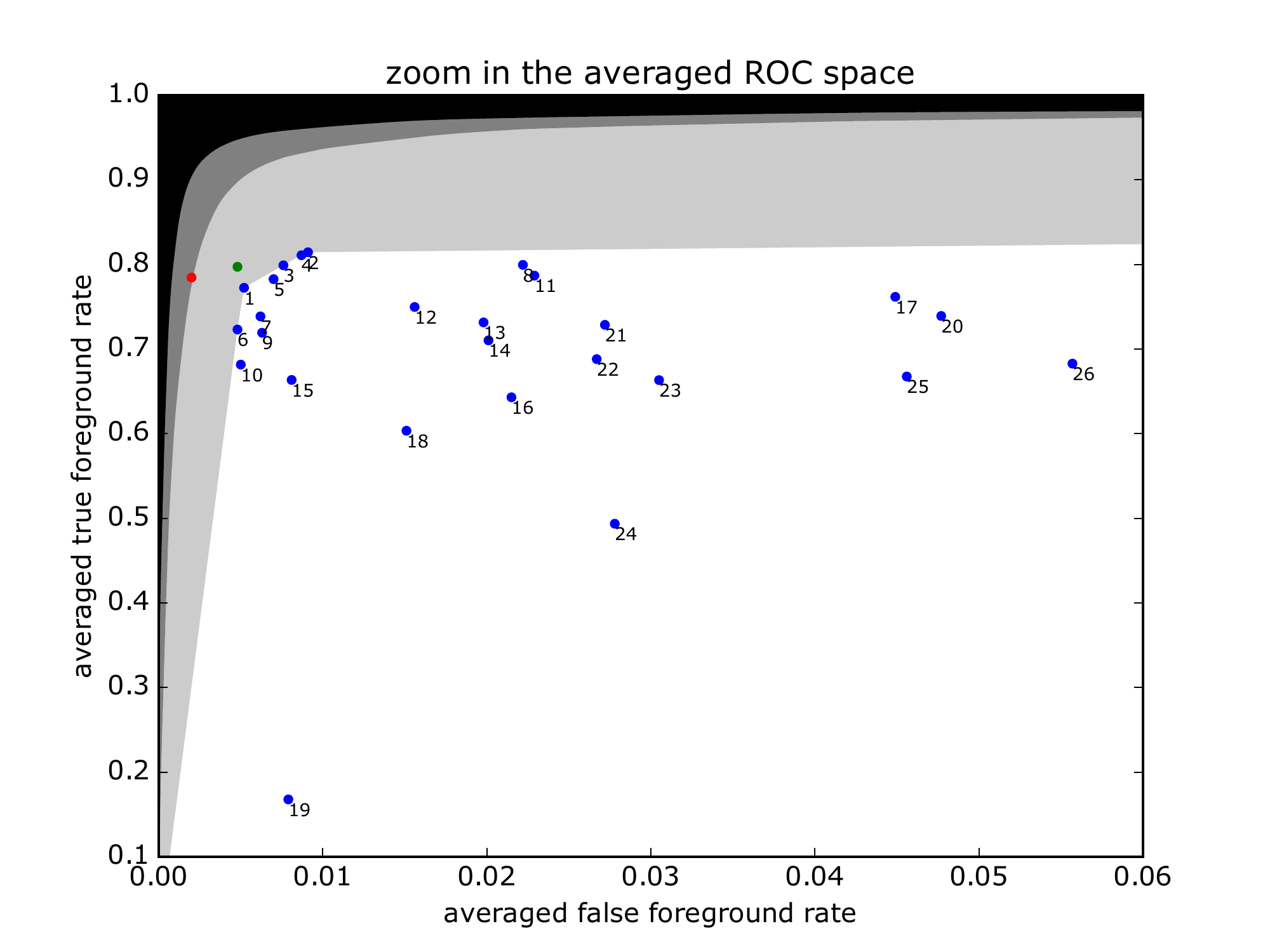}\caption{The $\overline{\textrm{ROC}}$ space with the $26$ $\protect\BackgroundSubtraction$
algorithms ($\initial$), previous state-of-the-art combination results
($\iutisfive$ for IUTIS-5~\cite{Bianco2017Combination} and \textcolor{cnnsfc}{$\bullet$}
for CNN-SFC~\cite{Zeng2018Combining}), and the sets of achievable
performances with the strategies \RandomChoice ($\randomselection$
for $n=26$, see Section~\ref{subsec:Technique-with-the-strategy-random-choice})
and \AllCombinations ($\maxnine$ for $n\le9$, and $\twentysix$
for $n=26$, see Section~\ref{subsec:Technique-with-the-strategy-all-combinations}).\label{fig:overall-ROC-1}}
\end{figure}
We analyze the sets of achievable performances for the $6$ combination
strategies and $26$ unsupervised $\BackgroundSubtraction$ algorithms.

\paragraph*{A huge potential for the pixelwise combinations.}

Figure~\ref{fig:overall-ROC-1} shows individual performances and
sets of achievable performances in $\overline{\textrm{ROC}}$. According
to it, the margin for improving the $\BackgroundSubtraction$ performance
is huge. Some pixelwise combinations of outputs (\AllCombinations)
can drastically outperform all the individual $\BackgroundSubtraction$
algorithms listed in Table~\ref{tab:Pool-of-combinable}. They can
also largely outperform the simple \RandomChoice strategy. Moreover,
there exist achievable performances that are closer to the oracle
(upper left corner) than those of the non-pixelwise combinations IUTIS-5
and CNN-SFC.

\paragraph*{There are efficient combination strategies.}

\noindent 
\begin{figure}[t]
\centering{}\includegraphics[width=0.9\columnwidth]{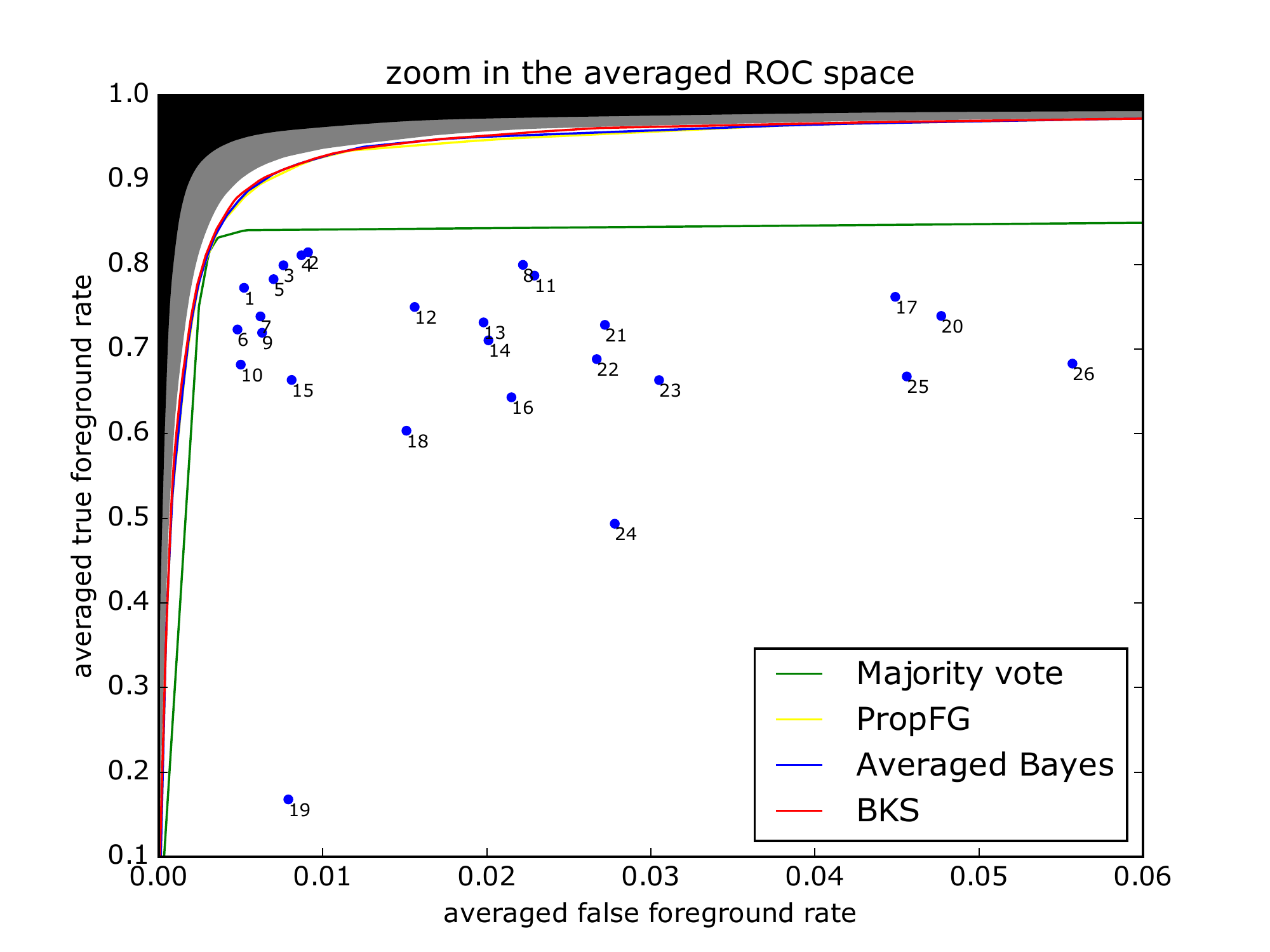}\caption{The $\overline{\textrm{ROC}}$ space with the convex hulls of performances
achievable by combining $n\le9$ algorithms with four deterministic
strategies (colored curves, see Section~\ref{subsec:Technique-with-other-strategies}).\label{fig:overall-ROC-2}}
\end{figure}
A combination strategy is efficient when (1) it produces a tractable
amount of combinations, and (2) most of the performances achievable
with \AllCombinations are also achievable by randomly choosing between
some of the combinations produced by that strategy. Table~\ref{tab:combination-strategies}
confirms that the amounts of combinations produced by our four deterministic
strategies are all largely inferior to $2^{2^{n}}$. Figure~\ref{fig:overall-ROC-2}
facilitates the comparison between the convex hulls of their performance
point clouds and the achievable zone with \AllCombinations, in $\overline{\textrm{ROC}}$.
We see that \PropFG, \AveragedBayes, and \BKS are efficient (and
have approximately the same convex hulls), but not the classical \MajorityVote.
Little has to be gained from other pixelwise combination strategies
(training decision trees or deep neural networks, adding some regularization
to \BKS, \ldots ) if the amount of combined algorithms is not increased.

\noindent 
\begin{figure}[t]
\begin{centering}
\textsf{\includegraphics[width=0.9\columnwidth]{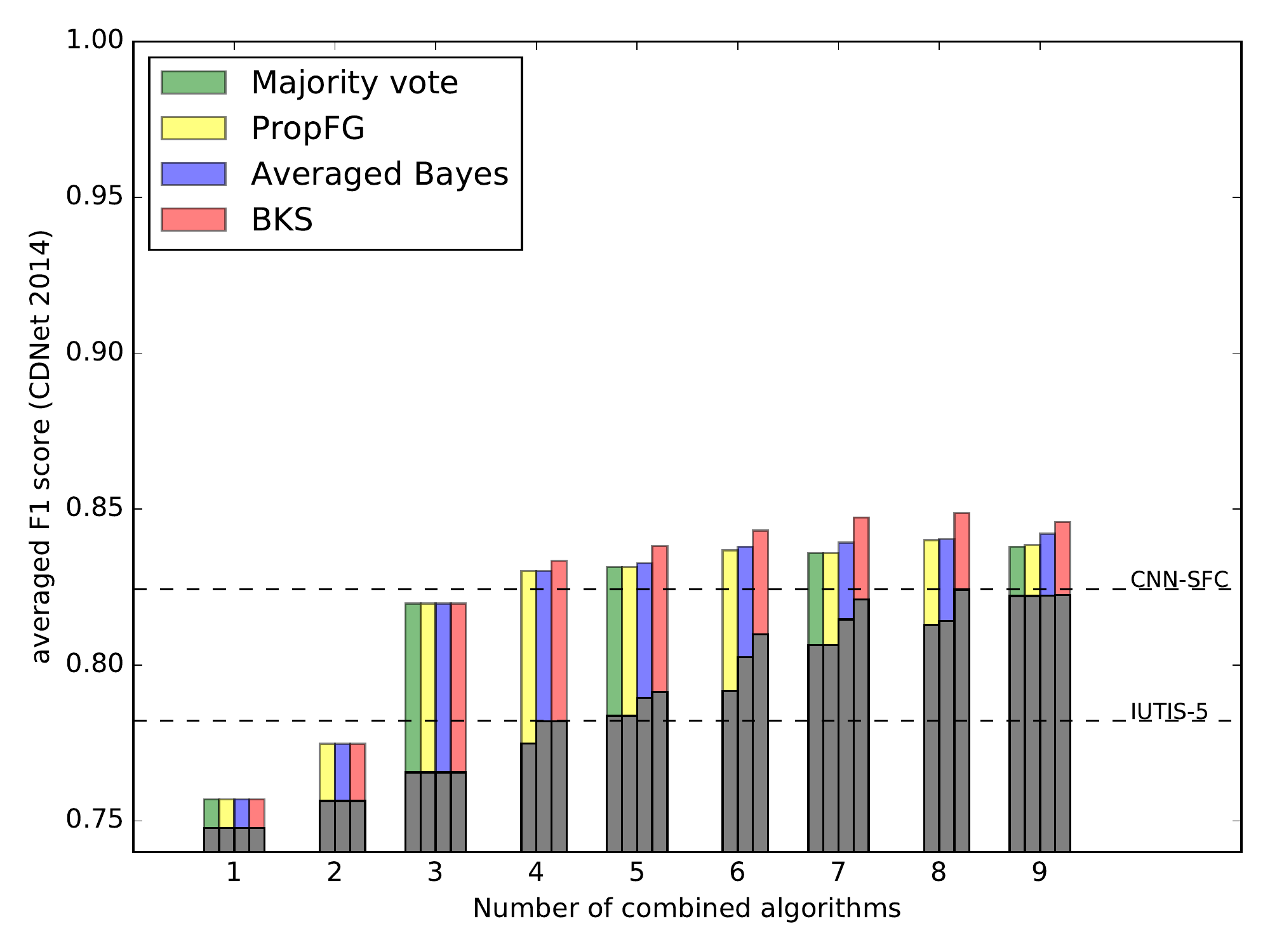}}
\par\end{centering}
\caption{Best $\overline{\textrm{F}_{1}}$ scores for four deterministic combination
strategies obtained with respect to the number $n=k,\,k\in\{1,\ldots,9\}$
of combined $\protect\BackgroundSubtraction$ algorithms. Gray bars
indicate the results obtained with the top-$n$ algorithms (ranking
of \CDNET) instead of an optimized selection of $n$ algorithms.\label{fig:Meilleurs-scores-F1}}
\end{figure}
\begin{table*}
\begin{centering}
\resizebox{\textwidth}{!}{
\begin{tabular}{|c|c|c|c|c|c|}
\hline 
 & Combination strategy & Selection of $\BackgroundSubtraction$ algorithms & $n$ & Threshold $\threshold$ & $\overline{\textrm{F}_{1}}$ score\tabularnewline
\hline 
\hline 
\multirow{2}{*}{Previous works} & IUTIS-5~\cite{Bianco2017Combination} & $\SuBSENSE$ + $\FTSG$ + $\CwisarDH$ + $\Spectral$ + $\AMBER$ & $5$ & None & $0.7821$\tabularnewline
\cline{2-6} \cline{3-6} \cline{4-6} \cline{5-6} \cline{6-6} 
 & CNN-SFC~\cite{Zeng2018Combining} & $\SuBSENSE$ + $\FTSG$ + $\CwisarDH$ & $3$ & Unknown & \textcolor{black}{$0.8243$}\tabularnewline
\hline 
\hline 
\multirow{4}{*}{Ours} & \MajorityVote & $\PAWCS$ + $\WeSamBE$ + $\FTSG$ + $\CwisarDRP$ + $\MBS$ + $\CEFIC$
+ $\MBSv$ + $\EFIC$ + $\GraphCutDiff$ & $9$ & None & $0.8380$\tabularnewline
\cline{2-6} \cline{3-6} \cline{4-6} \cline{5-6} \cline{6-6} 
 & \PropFG & $\PAWCS$ + $\WeSamBE$ + $\FTSG$ + $\CwisarDRP$ + $\MBS$ + $\CEFIC$
+ $\MBSv$ + $\EFIC$ & $8$ & $\nicefrac{7}{16}$ & $0.8401$\tabularnewline
\cline{2-6} \cline{3-6} \cline{4-6} \cline{5-6} \cline{6-6} 
 & \AveragedBayes & $\PAWCS$ + $\WeSamBE$ + $\FTSG$ + $\CwisarDRP$ + $\MBS$ + $\CEFIC$
+ $\MBSv$ + $\EFIC$ + $\GraphCutDiff$ & $9$ & $0.2918$ & $0.8421$\tabularnewline
\cline{2-6} \cline{3-6} \cline{4-6} \cline{5-6} \cline{6-6} 
 & \BKS & $\PAWCS$ + $\FTSG$ + $\CwisarDRP$ + $\MBS$ + $\CEFIC$ + $\MBSv$
+ $\EFIC$ + $\Euclidean$ & $8$ & $0.2720$ & \textbf{\textcolor{black}{$\mathbf{0.8487}$}}\tabularnewline
\hline 
\end{tabular}}
\par\end{centering}
\centering{}\caption{Summary of $\overline{\textrm{F}_{1}}$ scores including those of
previous works (non-pixelwise combinations), and the maximum scores
that are achievable with our four deterministic combination strategies
by tuning the selection of algorithms and the threshold (pixelwise
combinations). Note that our best score ($0.8487$) surpasses the
one of CNN-SFC.\label{tab:Meilleurs-scores-F1-table}}
\end{table*}

\paragraph*{It is worth investigating combinations of $\protect\BackgroundSubtraction$
algorithms.}

\noindent Figure~\ref{fig:Meilleurs-scores-F1} shows that the best
$\overline{\textrm{F}_{1}}$ scores are, for $n>1$, significantly
better than those obtained without combination ($n=1$). This suggests
that looking for efficient combinations of existing $\BackgroundSubtraction$
algorithms or developing new $\BackgroundSubtraction$ algorithms
complementary to the existing ones, even if not necessarily better,
might be more profitable than searching for the best algorithm. Despite
the fact that this conclusion was already drawn in~\cite{Jodoin2014Overview},
the $\BackgroundSubtraction$ community continues to propose hundreds
of new $\BackgroundSubtraction$ algorithms every year, the best of
which work barely better than the state of the art, without investigating
the contribution of the proposed algorithms when combined with those
already described in the literature.

\paragraph*{How should we combine?}

\noindent Figure~\ref{fig:Meilleurs-scores-F1} also helps in observing
that our four deterministic strategies achieve the same maximal $\overline{\textrm{F}_{1}}$
score for $n\in\left\{ 2,3\right\} $. For $4\leq n\leq9$, the ranking
according to $\overline{\textrm{F}_{1}}$ is: $\MajorityVote\leq\PropFG\simeq\AveragedBayes\leq\BKS$.
Despite that, the improvement of \AveragedBayes and \BKS performance
is too small to balance their much larger amount of combinations to
test in practice. 

\paragraph*{What should we combine?}

For any given $n$, it might be tempting to combine the top-$n$ algorithms.
In fact, Figure~\ref{fig:Meilleurs-scores-F1} shows that we can
do much better by carefully cherry picking the combined algorithms
among all the available ones (see colored bars vs. gray bars). Moreover,
the difference in performance between a colored bar and the corresponding
gray bar is, in most cases, greater than the difference in performance
between adjacent gray bars. This suggests that knowing precisely what
to combine is more important than knowing precisely how to combine.
Poorly ranked algorithms can be useful when combined with others,
even if they do not perform well alone. This is illustrated in Table~\ref{tab:Meilleurs-scores-F1-table},
where we can observe that our best result are obtained by selecting
algorithms in different zones of the leaderboard.

\paragraph*{A new \textquotedblleft state-of-the-art\textquotedblright{} $\textrm{F}_{1}$
score on \CDNET 2014.}

Our four deterministic strategies can outperform IUTIS-5 and CNN-SFC
when the combined algorithms and the threshold $\threshold$ are adequately
chosen. As shown in Table~\ref{tab:Meilleurs-scores-F1-table}, our
results establish a new ``state-of-the-art'' $\textrm{F}_{1}$ score
of $0.8487$ on \CDNET 2014, against $0.8243$ for the previous one.

\section{Conclusion}

\label{sec:Conclusion}

To push the performance in $\BackgroundSubtraction$, one can either
develop new algorithms and publish their results on \CDNET, or develop
combinations based on the results already available on this platform.
Our results show that such combinations have the potential to outperform
the individual algorithms. This has resulted in six conclusions. Our
findings were all made possible thanks to the availability of outputs
on the \CDNET platform, a choice that should be promoted for all
challenges!

\paragraph*{Acknowledgment. }

\noindent This work was supported by the Service Public de Wallonie
(SPW) Recherche, under Grant $\text{N}^{\text{o}}$.~2010235 --
ARIAC by \href{https://DigitalWallonia4.ai}{DigitalWallonia4.ai}.

\end{document}